\documentclass[manuscript,screen]{acmart}
\usepackage[lined]{algorithm2e}
\usepackage{amsmath}
\usepackage{array}
\usepackage{bm}
\usepackage{appendix}

 \usepackage{multirow} 
\AtBeginDocument{%
  \providecommand\BibTeX{{%
    \normalfont B\kern-0.5em{\scshape i\kern-0.25em b}\kern-0.8em\TeX}}}

\setcopyright{acmcopyright}
\copyrightyear{2018}
\acmYear{2024}
\acmDOI{XXXXXXX.XXXXXXX}
\acmBooktitle{Woodstock '18: ACM Symposium on Neural Gaze Detection,
 June 03--05, 2018, Woodstock, NY} 
\acmPrice{15.00}
\acmISBN{978-1-4503-XXXX-X/18/06}

\begin{document}

\title{An Unbiased Risk Estimator for Partial Label Learning with
Augmented Classes}

\author{Jiayu Hu}
\email{hujiayu@cqu.edu.cn}
\orcid{0000-0003-2423-9465}
\author{Senlin Shu}
\email{shusenlin@126.com}
\author{Beibei Li}
\authornote{Corresponding author.}
\email{libeibeics@cqu.edu.cn}
\orcid{0000-0003-2711-9370}
\author{Tao Xiang}
\email{txiang@cqu.edu.cn}
\orcid{0000-0002-9439-4623}
\author{Zhongshi He}
\email{zshe@cqu.edu.cn}
\affiliation{%
  \institution{College of Computer Science, Chongqing University}
  \city{Shapingba District}
  \state{Chongqing}
  \country{China}
}

\renewcommand{\shortauthors}{Hu, et al.}

\begin{abstract}

 Partial Label Learning (PLL) is a typical weakly supervised learning task, which assumes each training instance is annotated with a set of candidate labels containing the ground-truth label. Recent PLL methods adopt identification-based disambiguation to alleviate the influence of false positive labels and achieve promising performance. However, they require all classes in the test set to have appeared in the training set, ignoring the fact that new classes will keep emerging in real applications. To address this issue, in this paper, we focus on the problem of Partial Label Learning with Augmented Class (PLLAC), where one or more augmented classes are not visible in the training stage but appear in the inference stage. Specifically, we propose an unbiased risk estimator with theoretical guarantees for PLLAC, which estimates the distribution of augmented classes by differentiating the distribution of known classes from unlabeled data and can be equipped with arbitrary PLL loss functions. Besides, we provide a theoretical analysis of the estimation error bound of the estimator, which guarantees the convergence of the empirical risk minimizer to the true risk minimizer as the number of training data tends to infinity. Furthermore, we add a risk-penalty regularization term in the optimization objective to alleviate the influence of the over-fitting issue caused by negative empirical risk. Extensive experiments on benchmark, UCI and real-world datasets demonstrate the effectiveness of the proposed approach.

\end{abstract}






\maketitle

\section{Introduction}
Supervised learning models have been vigorously developed over the past few years \cite{TIST3}. Although having achieved promising performance, they rely on a large number of accurately labeled instances to complete training, which is not only costly but also suffers from difficulties in data acquisition caused by privacy and security issues. Weakly supervised learning \cite{weak_supervise}, which utilizes incomplete labels, inaccurate labels and inexact labels to train models, has drawn extensive attention in recent years. Several representative tasks have been investigated, such as semi-supervised learning \cite{zhu2022introduction,chapelle2009semi}, noisy-label learning \cite{nosiylabel,noisylabel2}, partial-label learning \cite{xie2018partial}, multi-label learning \cite{multilabel1,multilabel2,active1,active2,TIST2}, etc. They have been widely employed in various real-world scenarios, including data annotation \cite{raykar2010learning}, disease diagnosis \cite{TIST1}, object segmentation \cite{TIST4,TIST5}, object detection \cite{han2024progressive} and text classification \cite{TIST6}.  
\begin{figure}[t]

\includegraphics[scale=0.47]{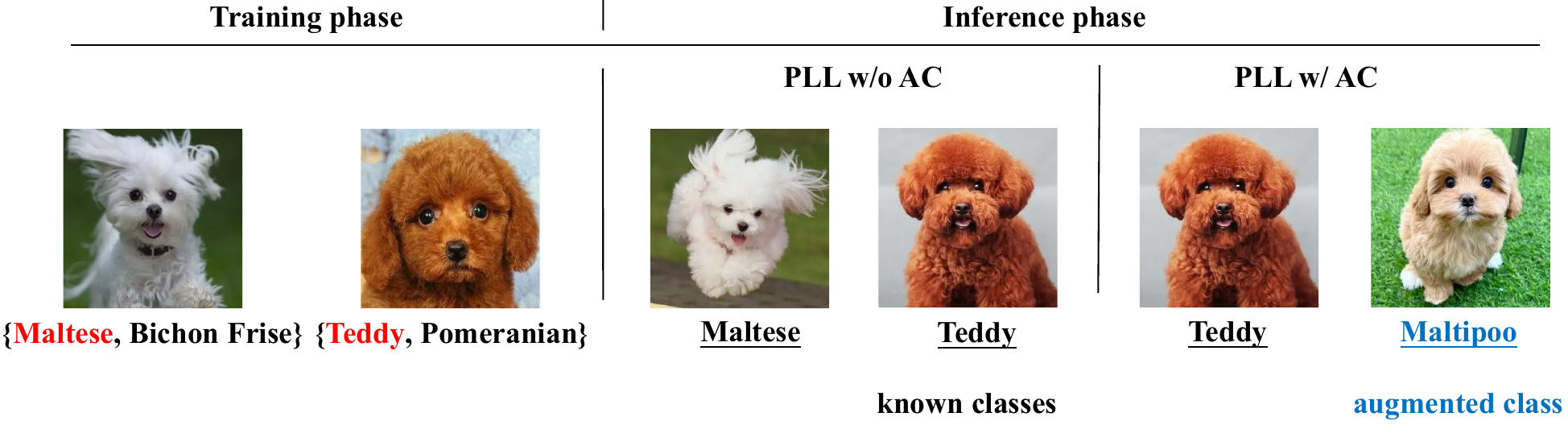}
\caption{An comparison example between PLL without Augmented Classes and PLL with Augmented Classes, where  all the classes in test set are known when PLL without Augmented Classes while augmented classes emerges in test set when PLL with Augmented Classes.}

\label{example}
\end{figure}

Partial Label Learning  (PLL), a typical weakly supervised learning task,  assumes each training instance is annotated with a candidate label set containing its ground-truth label. For example, as shown in Figure \ref{example}, the annotator cannot clearly distinguish whether the dog in the first picture is Maltese or Bichon Frise due to their  appearance similarity, so he/she annotates the picture with a candidate label set $\{Maltese, Bichon\ Frise
\}$. Partially labeled data are ubiquitous, easy to collect, and large in quantity. Therefore, PLL has been widely applied in various practical applications, such as automatic image annotation  \cite{chen_learning_2018} and multimedia content analysis  \cite{zeng_learning_2013}.

The largest challenge of PLL is label ambiguity, that is, false positive labels in candidate label sets could mislead the model during the training phase. Recent methods \cite{RCCC,PRODEN,DPLL,PiCO} mainly leverage identification-based disambiguation to resolve label ambiguity. They try to gradually identify the ground-truth label of each instance during model training, so as to reduce the influence of false positive labels. However, existing PLL models require the classes in the test set have all appeared in the training set, which may not be guaranteed in practice. In real-world scenarios, new classes of instances keep emerging, and there will be one or more augmented classes that are not visible in the training stage but appear in the inference stage. For example, as shown in Figure \ref{example}, the augmented class Maltipoo does not appear in the training stage. Partial Label Learning with Augmented Class (PLLAC) requires us to learn a classifier that not only accurately classifies known classes but also effectively recognizes the augmented classes. 

In recent years, several methods \cite{LAC3,LAC4} towards Learning with Augmented Classes  (LAC)  have been proposed by exploiting the relationship between known and augmented classes. However, they only work for cases where accurate labels are available. PLLAC, which is a harder and more pervasive problem, has not been investigated.

Zero-shot learning, a type of transfer learning, can classify unseen classes and has been extensively studied\cite{yan2022semantics,liu2023simple}. However, it relies on semantic auxiliary information about the classes and requires supervised training samples with accurate and unique labels, which is not always available. Unlike zero-shot learning, in the PLLAC setting, the training samples are not fully supervised. Motivated by that the distribution of augmented classes can be estimated by differentiating the distribution of known classes from unlabeled test data, we exploit unlabeled data to facilitate PLLAC. Our contributions are summarized as follows:

\begin{itemize}
\item  (\textbf{Method}) We propose a generalized unbiased risk estimator with theoretical guarantees for PLLAC, which exploits unlabeled data to estimate the distribution of augmented classes and is divided into PLL part and unlabeled part. PLL part can be equipped with an arbitrary PLL loss functions.
\item  (\textbf{Theory})   We derive an estimation error bound for the estimator, which guarantees that the obtained empirical risk minimizer would approximately converge to the expected risk minimizer as the number of training data tends to infinity.
\item  (\textbf{Experiments})   We conduct extensive experiments on both benchmark datasets and real-world datasets to demonstrate the effectiveness of the proposed estimator.
\end{itemize}

The rest of the paper is organized as follows. Section \ref{sec:relatedwork} is about the related work of the PLLAC problem. We propose our method and give theoretical analysis in Section \ref{sec:proposedmethod}. We describe the experimental setting details and report experiment results in Section \ref{sec:exp}, where extensive experiments are conducted to demonstrate the effectiveness of our method. Finally, we conclude the paper.

\section{RELATED WORK}\label{sec:relatedwork}

\subsection{Partial Label Learning}
Existing PLL methods adopt label disambiguation to mitigate the influence of label ambiguity on model training. They can be roughly divided into average-based methods \cite{ADS2} and identification-based methods \cite{IDS1,IDS2} according to different disambiguation strategies. Average-based methods treat each candidate label of training instances equally and make prediction by averaging the outputs on each candidate label. Though simple to implement, they make the ground-truth label overwhelmed by false positive labels easily. Identification-based methods try to identify the ground-truth label from the candidate label set, so as to reduce the influence of false positive labels. Some of them adopt two-phase strategy \cite{zhang_solving_2015}, i.e., first refine label confidence, then learn the classifier, while others progressively refine confidence during learning the classifier \cite{yu_maximum_2016}. Early PLL methods are usually linear or kernel-based models, which are hard to deal with large-scale datasets. With the powerful modeling capability and the rapid development of deep learning, deep PLL methods, which can handle high-dimensional features, have drawn attentions in recent years. Most deep PLL models are identification-based methods, for example, RC \cite{RCCC}, PRODEN \cite{PRODEN} and LWS  \cite{LWS} estimate label confidence and train the model with it iteratively. Furthermore, PiCO \cite{PiCO} and DPLL \cite{DPLL}  explore contrastive learning and manifold consistency in deep PLL, respectively.  However, existing PLL methods ignore the fact that new labels may emerge during the inference process in practice and are not able to deal with the  augmented classes in partial label learning. 

\subsection{Open-Set Recognition}
Open-set recognition (OSR) considers a more realistic scenario, where incomplete knowledge of the world exists at training time and unknown classes at test time appear. \cite{scheirer2012toward}. Studies for OSR problem could divided into two categories: discriminative models and generative models \cite{geng2020recent}. Most methods based on discriminative model study the relationship between known classes and augmented classes, like using open space risk based on SVM \cite{scheirer2014probability} as the traditional ML-based methods, a Nearest Non-Outlier (NNO) algorithm \cite{bendale2015towards} is established for open-set recognition by using the Nearest Class Mean (NCM) classifier \cite{mensink2013distance} as distance-based method. The way based on DNN is to exploit convolutional neural network which applies a threshold on the output probability, that is OpenMax\cite{Bendale_2016_CVPR}, using an alternative for the SoftMax function as the final layer of the network \cite{bendale2016towards}. Learning with Augmented classes is similar to OSR problem in Pattern Recognition for they both deal with the problem that to classify the augmented classes which are unseen in training stage but emerge in test phase. Different from OSR problem, LAC studies on how to classify all the classes appeared in test stage, while OSR focuses on whether observed instances are out of distribution.

\subsection{Learning with Augmented Classes}


Learning with augmented classes (LAC) is a problem where augmented classes unobserved in the training stage may emerge in the test phase. It is a main task of class-incremental learning \cite{zhou2002hybrid, tao2020few}. The main challenge of LAC is that no instances from augmented classes appear in the training phase. Motivated by that unlabeled data can be easily collected in real-world application and unlabeled data help improve the classification performance when the number of training instances is limited \cite{zhou2010semi,zhu2022introduction,semisupervise}, Da et al. \cite{LAC3} present the LACU (Learning with Augmented Class with Unlabeled data) framework and the LACU-SVM approach to the learning with augmented class problem. Considering the distribution information of augmented classes may be contained in unlabeled data and estimated by differentiating the distribution of known classes from unlabeled data, a recent study proposes an unbiased risk estimator (URE) under \emph{class shift condition} \cite{LAC4}, which exploits unlabeled data drawn from test distribution. However, this
URE is only restricted to the specific type of one-versus-rest loss functions for multi-class classification. Therefore, Shu et al. \cite{LAC5} propose a Generalized Unbiased Risk Estimator (GURE) which can be equipped with arbitrary loss functions and provide a theoretical analysis on the estimation error bound. Under the assumption that the distribution of known classes would not change when augmented classes emerged in test phase, both URE and GURE introduce the class shift condition to depict the relationship between known and augmented class, then the testing distribution $p_{\text{te}}$ can be obtained as:
\begin{align}
\label{test_distribution}
     P_{\text{te}} = \theta\cdot P_{\text{kc}}   +  (1-\theta) \cdot P_{\text{ac}} 
\end{align}
where $\theta \in \left[0,1\right]$ is a mixture proportion of the distribution of known classes $P_{\text{kc}}$ and augmented classes $P_{\text{ac}}$.

However, these methods for the LAC problem are all towards supervised learning and not applicable to partially labeled datasets.

\section{THE PROPOSED METHOD}\label{sec:proposedmethod}
In this section, we first present the formulation of the PLLAC problem. Next, we propose an unbiased risk estimator for the PLLAC problem and provide theoretical analysis for it. Then, we identify the potential over-fitting issue of unbiased risk estimator and establish a risk-penalty regularization to alleviate the over-fitting problem.

\subsection{Problem Formulation}

We represent the feature space and label space of partial label data respectively as $\mathcal{X},\mathcal{Y}$, where $\mathcal{X}\in\mathbb{R}^d$ and $\mathcal{Y} = \{1,\dots,k\}$, $d$ is feature dimension and $k$ is the number of classes. In conventional PLL, we are given a $k$-classes PLL dataset $\mathcal{D}_{\mathrm{PL}} = \{\boldsymbol{x}_i, S_i\}_{i=1}^{n}$ independently and identically drawn from an underlying distribution with probability density $P_{\mathrm{PL}}$ defined over $\mathcal{X}\times \mathcal{Y}$, where each training sample $\boldsymbol{x}_i$ is associated with a candidate label set $S_i$, $S_i\in\mathcal{C}$, and $\mathcal{C} = \{2^{\mathcal{Y}}\setminus \varnothing \setminus \mathcal{Y}\}$. The goal of PLL is to train a classifier $f:\mathcal{X}\rightarrow \mathcal{Y}$ with considering the training set and test set are under the same data distribution. However, in the test phase of the PLLAC, augmented classes unobserved in the training phase may emerge. Due to uncertainty and inaccessibility of the number of augmented classes in test set, these augmented classes would be labeled as one class named $ac$ and the augmented label space could be denoted as $\mathcal{Y}^{\prime}=\{1,\dots,k,ac\}$.

In addition, we assume that in the training stage, except for the partially labeled data, a set of unlabeled data sampled from the test set, denoted as $D_\mathrm{U}=\left\{\boldsymbol{x}_i\right\}_{i=1}^{n_\mathrm{U}}\sim p_{\text{te}} (\boldsymbol{x},y)  $, is available and could be used in training stage. Note that it is feasible to use unlabeled data from test set when training the model, because in most situations, it is easy to obtain test set or open-set data with the same distribution as the test set. The unlabeled data enrich the features of training instances without supervision leakage, thus are able to improve the generalization ability of the model in the case of distribution drift.

Therefore, the goal of PLLAC is to learn a $k+1$ multi-class classifier based on partially labeled data and unlabeled data sampled from test set distributions, which can obtain the minimal empirical risk in test set. The notations are listed in Table \ref{notation}.
\begin{table}[t]
    \centering
    \caption{Notations}
    \label{notation}
    \begin{tabular}{c|l}
    \toprule  
        Symbol& Description \\ \midrule
        $p_{\mathrm{te}},p_{\mathrm{ac}},p_{\mathrm{kc}}$& distribution of test dataset, augmented classes and known classes.\\
        $\theta$ & a mixture proportion of distribution of known classes and augmented classes. \\
        $S$& candidate label set.\\
        $\mathcal{C}$& a set that contains all possible candidate label set.\\
        $p_{ij}$& the confidence of the $i$-th sample, $j$-th class.\\
        $n,m,n_{\mathrm{U}}$& the number of training instances, testing instances and unlabeled instances.\\
        $f(x)$ & the calssification probability of instance $x$ in $k+1$ classes. \\
        $\ell(f(x),j)$& the loss on sample $i$ when given label $j$.\\
        $ \Omega(f)$& the loss for generalized risk-penalty regularization. \\
        $R_{\mathrm{un}},R_{\mathrm{reg}}$& expected risk for unbiased estimator and unbiased estimator with regularization term. \\
        $\widehat{R}_{\mathrm{un}},\widehat{R}_{\mathrm{reg}}$ & empirical risk for unbiased estimator and unbiased estimator with regularization term.\\
        $\lambda$& the weight of risk-penalty regularization term in $R_{\mathrm{Reg}}$.\\
        \bottomrule
    \end{tabular}
\end{table}
\subsection{Unbiased Risk Estimator}\label{sec:URE}

Similar to LAC, in PLLAC, the distribution of data from the augmented classes is also inaccessible. Therefore, we follow the \emph{class shift condition} \cite{LAC4} in Eq.  \ref{test_distribution} to describe the relationship between the distribution of known classes and augmented classes.  Specifically, on accurately labeled datasets, the distribution of known classes can be calculated by $P_{\text{kc}}=\sum_{j=1}^k p(\boldsymbol{x},y=j)$. However, in PLLAC, only partial labels are available and the distribution is calculated by  $P_{\text{PL}}=\sum_{v=1}^{|C|}p(\boldsymbol{x},S=S_v)$. Fortunately, we find that the two are equivalent by the following derivation,
\begin{align}
\label{eq:p_kcp_pl}
    P_{\text{kc}}=\sum_{j=1}^k p(\boldsymbol{x},y=j)=\sum_{v=1}^{|C|}\sum_{j=1}^kp(y=j|\boldsymbol{x},S_{v})p(\boldsymbol{x},S=S_v) =\sum_{v=1}^{|C|}p(\boldsymbol{x},S=S_v)=P_{\text{PL}},
\end{align}
where $p(y=j|\boldsymbol{x},S_{v})$ indicates the probability that $j$ is the true label with the given data $(\boldsymbol{x},S_{v})$ and $\sum_{j=1}^{k}p(y=j|\boldsymbol{x},S_{v})=1$. Therefore, we obtain the following distribution relationship in PLLAC by substituting  $P_{\text{kc}}$ with $P_{\text{PL}}$:
\begin{align}
\label{distribution}
    P_{\mathrm{te}} = \theta\cdot P_{\text{PL}}   +  (1-\theta)\cdot P_{\mathrm{ac}}.
\end{align}

Let $f (\boldsymbol{x})  \in\mathbb{R}^{\mathrm{k}+1}$ denote the classification probability of instance  $\boldsymbol{x}$ in $k+1$ classes, $\ell_{\mathrm{PLL}} (\cdot)  $ represents a PLL loss function, $\ell (\cdot) $ is multi-class classification loss function, e.g., the categorical cross-entropy loss. The loss of instance $\boldsymbol{x}$ in partial label set $S$ and \textit{ac} class can be represented respectively as $\ell_{\mathrm{PLL}} (f (\boldsymbol{x})  ,S)  $ and $\ell (f (\boldsymbol{x})  ,\mathrm{ac})  $. According to Eq. (\ref{distribution}), the risk estimator of the PLLAC problem over test set can be defined as:
\begin{align}
\label{risk1}
R (f)   &= \theta \mathbb{E}_{ (\boldsymbol{x},S)\sim P_{\mathrm{PL}} }[\ell_{\mathrm{PLL}} (f (\boldsymbol{x})  ,S)  ] +  (1-\theta)   \mathbb{E}_{x\sim P_{\mathrm{ac}}}  [\ell (f (\boldsymbol{x})  ,\mathrm{ac})  ].
\end{align}

Since $P_{\mathrm{ac}}$ is unknown in test set, $\mathbb{E}_{p_{\mathrm{ac}}(\boldsymbol{x})}[\ell (f (\boldsymbol{x})  ,\mathrm{ac})  ]$ cannot be calculated directly. Then, we need to permute it in another way. From Eq. (\ref{distribution}), we can obtain:
\begin{align}
\label{transformation}
     (1-\theta)  P_{\mathrm{ac}} = P_{\mathrm{te}} - \theta P_{\mathrm{PL}}.
\end{align}

Then, we calculate the expected risk on the \textit{ac} class for each side of the equation as following equation:
\begin{align}
\label{expected}
 (1-\theta)   \mathbb{E}_{x\sim P_{\mathrm{ac}}}   [\ell (f (\boldsymbol{x})  ,\mathrm{ac})  ] = \mathbb{E}_{x\sim P_{\mathrm{te}}}   [\ell (f (\boldsymbol{x})  ,\mathrm{ac})  ]- \theta \mathbb{E}_{(\boldsymbol{x},S)\sim P_{\mathrm{PL}}}[\ell (f (\boldsymbol{x})  ,\mathrm{ac})  ].
\end{align}

By substituting Eq. (\ref{expected}) into the expected risk Eq. (\ref{risk1}). We can obtain:
\begin{align}
\label{unbiased_risk}
    R_{\mathrm{un}} (f)   &= \theta \mathbb{E}_{ (\boldsymbol{x},S)\sim P_{\mathrm{PL}}}[\ell_{\mathrm{PLL}} (f (\boldsymbol{x})  ,S)  ] + \mathbb{E}_{x\sim P_{\mathrm{te}}   }[\ell (f (\boldsymbol{x})  ,\mathrm{ac})  ]- \theta \mathbb{E}_{(\boldsymbol{x},S)\sim P_{\mathrm{PL}}}[\ell (f (\boldsymbol{x})  ,\mathrm{ac})  ],
\end{align}
which is an unbiased risk estimator. The $\ell_{\mathrm{PLL}}$ could be an arbitrary PLL loss. Therefore, we can learn a classifier from unlabeled data in test distribution and partially labeled data in training set via this estimator. 

Given $n$ partially labeled instances in training set, that is, $D_{\mathrm{PLL}}=\left\{\boldsymbol{x}_i,S_i\right\}_{i=1}^n$, and $n_{\mathrm{U}}$ unlabeled instances sampled from test set $D_{\mathrm{U}}=\left\{\boldsymbol{x}_i\right\}_{i=1}^{n_{\mathrm{U}}}$, we can calculate its empirical risk estimator which is approximate to the expected risk $R_{\mathrm{un}} (f)  $. We employ the softmax function in the last layer of classifier $f (\cdot)$ to calculate classification probability of total $k+1$ classes. In fact, the loss  $\ell_{\mathrm{PLL}}$  could be arbitrary partial-label learning loss, e.g. CC \cite{RCCC}, PRODEN \cite{PRODEN}, RC \cite{RCCC}. We conduct a series of experiments in Section \ref{sec:loss} and find out that performance RC outperforms others. 
Therefore, we choose RC to instantiate $\ell_{\mathrm{PLL}}(\cdot)$. Besides, we utilize cross entropy to calculate $\ell (f (\boldsymbol{x}), \mathrm{ac}) $. Then, we can obtain the empirical approximation of $R_{\mathrm{un}}(f)$: 
\begin{align}
\label{empirical_unbiased_risk}
\widehat{R}_{\mathrm{un}} (f)   = \theta\frac{1}{n}\sum\limits_{i=1}^n\sum\limits_{j=1}^k {p}_{ij}\cdot \mathcal{L}(f(\boldsymbol{x}_i),j)  + \frac{1}{n_{\mathrm{U}}}\sum\limits_{i=1}^{n_{\mathrm{U}}} -\log f_{\mathrm{ac}} (\boldsymbol{x}_i)  + \theta\frac{1}{n}\sum\limits_{i=1}^n \log f_{\mathtt{ac}} (\boldsymbol{x}_i),
\end{align}
where ${p}_{ij}$ indicates the confidence of the $j$-th label be the true label of the $i$-th sample, $ \text{if } j \in S_i, {p}_{ij}= \frac{p (y_{i}=j \mid \boldsymbol{x}_{i})  }{\sum_{o \in S_{i}} p (y_{i}=o \mid \boldsymbol{x}_{i}) }$, $ \text{else } {p}_{ij}= 0$. Same to RC \cite{RCCC},   we estimate $p (y_{i}=j \mid \boldsymbol{x}_{i})$ by the  classification probability calculated by the model in the last epoch. $\mathcal{L}(f(\boldsymbol{x}_i),j)$ is the loss function for calculating the loss on sample $\boldsymbol{x}_i$ when given label $j$ and $\mathcal{L}(f(\boldsymbol{x}_i),j) = -\log f_j (\boldsymbol{x}_i)$ here, $f_j (\boldsymbol{x}_i)$ is the $j$-th element of $f (\boldsymbol{x}_i) $. Therefore, by minimizing $\widehat{R}_{\mathrm{un}} (f)  $ we can learn an effective classifier for partial-label learning with augmented classes. 

It is obvious that minimizing $\widehat{R}_{\mathrm{un}} (f)  $  would require estimating mixture proportion $\theta$. And we employ the Kernel Mean Embedding (KME)-based algorithm \cite{ramaswamy2016mixture}, which could obtain an estimator $\widehat{\theta}$ that would converge to true proportion $\theta$ under the separability condition, given unlabeled data and labeled data in training stage. 


\subsection{Theoretical Analysis}\label{sec:theoretical}
Here, we establish  an estimation error bound for our proposed unbiased risk estimator and prove that the estimator is consistent.\\
\textbf{Definition 1}  (Rademacher Complexity \cite{bartlett2002rademacher})   \emph{Let $n$ be a positive integer, $\boldsymbol{x}_1,...,\boldsymbol{x}_n$ be independent and identically distributed random variables drawn from a probability distribution with density $\mu$, $\mathcal{F}=\{f:\mathcal{X} \mapsto \mathbb{R}\}$ be a class of measurable functions. Then the expected Rademacher complexity of $\mathcal{F}$ is defined as}
\begin{align}
  \mathfrak{R}_n (\mathcal{F})  =\mathbb{E}_{\boldsymbol{x}_1,...,\boldsymbol{x}_n\sim\mu}\mathbb{E}_{\sigma}\left[\mathop{\mathrm{sup}}\limits_{f\in\mathcal{F}}\frac{1}{n}\sum\limits_{i=1}^n \sigma_{i}f (\boldsymbol{x}_i)  \right],
\end{align}
\emph{where $\boldsymbol{\sigma}= (\sigma_1,...,\sigma_n)  $ are Rademacher variables taking the value from} \{-1,+1\} \emph{with even probabilities.}

We denote by $\mathfrak{R}_n (\mathcal{F}_y)  $ the Rademacher complexity of $\mathcal{F}_y$ for the $y$-th class, where $\mathcal{F}_i=\{\boldsymbol{x}\mapsto f_i (\boldsymbol{x})  |f\in\mathcal{F}\}$. It is not hard to know that for all $y \in \mathcal{Y}$, $\mathfrak{R}_n (\mathcal{F}_y)  \leq C_{\mathcal{F}}\/ \sqrt{n}$, where $C_{\mathcal{F}}$ is a positive constant.

Let $\widehat{f}_{\mathrm{un}}=\arg\min_{f\in\mathcal{F}}\widehat{R}_{\mathrm{un}} (f)  $ be the empirical unbiased risk minimizer, and $f^*=\arg\min_{f\in\mathcal{F}}R (f)  $ be the true risk minimizer, then we have following theorem.\\
\textbf{Theorem 1.} \emph{Assume the loss function $\mathcal{L} (f (\boldsymbol{x})  , y)  $ is $\rho$-Lipschitz with respect to $f (\boldsymbol{x})   (0<\rho<\infty)  $ for all $y \in \mathcal{Y}$ and upper-bounded by $C_{\mathcal{L}}$, i.e., $C_{\mathcal{L}} = \sup_{x\in\mathcal{X},f\in\mathcal{F},y\in\mathcal{Y}}\mathcal{L} (f (\boldsymbol{x})  ,y)  $. Then, for any $\delta>0$, with probability at least  $1-\delta$,}
\begin{align}
\label{unbiased}
    R (\widehat{f}_{\mathrm{un}})  -R (f^*)  \leq C_{k,\rho,\delta}\left (\frac{3\theta}{2\sqrt{n}}+\frac{1}{\sqrt{m}}\right),  
\end{align}
where $C_{k,\rho,\delta}=\left (4\sqrt{2}\rho (k+1)  C_{\mathcal{F}}+C_{\mathcal{L}}\sqrt{\frac{\mathrm{log}\frac{2}{\delta}}{2}}\right)  $.

The proof of Theorem 1 is provided in Appendix. Therefore, we prove that the method is consistent, which means the  empirical risk minimizer $f_{un}$ would converge to the true risk minimizer $f^*$ as $m,n\rightarrow\infty$. 

\subsection{Overfitting of Unbiased Risk Estimator}
\begin{figure}[htb]

 \centering
\includegraphics[scale=0.32]{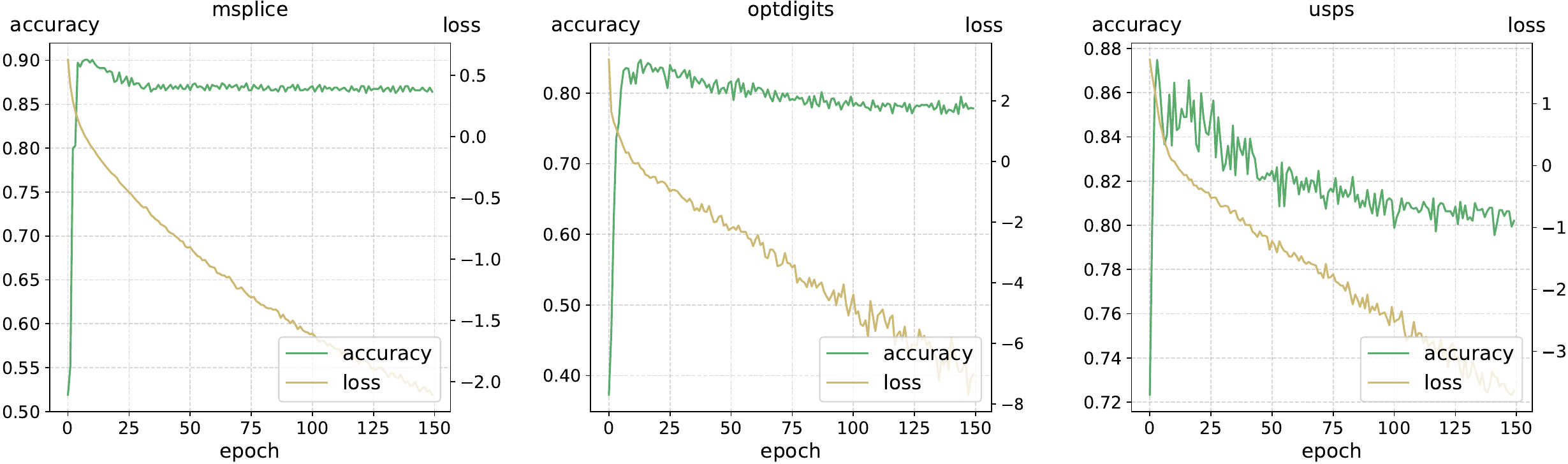}
\caption{Test performance on UCI datasets using $\widehat{f}_{\text{un}}$ in the training stage}

 \label{overfitting}
\end{figure}
Considering the classification loss on class \textit{ac} we used in the experiment is cross-entropy loss, which is unbounded above, the third term in the URE $\widehat{R}_{\mathrm{un}} (f)  $ could be unbounded below. Then, during training,  the loss in the training stage would steadily decrease and cause over-fitting issue. As shown in Figure \ref{overfitting}, during the first 20 epochs, the training loss decreases but does not fall below 0, and the accuracy on the training set increases accordingly. However, when the number of epochs increases, the training loss does not converge after it decreases below 0, resulting in the occurrence of overfitting issue, and the corresponding accuracy declines.

Previous work solves the overfitting problem by regularization \cite{LAC5}, motivated by this, we add a regularization term to alleviate the influence of the negative empirical risk. Notice that in $\widehat{R}_{\text{un}}$, the second and third term are both in the right side of Eq.~(\ref{expected}). Meanwhile, the left side is non-negative due to $1-\theta$ and $\ell (f (\boldsymbol{x})  ,\mathrm{ac})$ would not be below 0. Therefore, we could choose these two terms as $\widehat{R}_{\text{PAC}}$, which should also be non-negative  to be the regularization term. That is:

\begin{align}
    \widehat{R}_{\mathrm{PAC}}=\frac{1}{n_\mathrm{U}}\sum\limits_{j=1}^{n_\mathrm{U}} \ell\left (f (\boldsymbol{x}_j)  ,\mathrm{ac}\right)  -\frac{\theta}{n}\sum\limits_{i=1}^n\ell\left (f (\boldsymbol{x}_i)  ,\mathrm{ac}\right).
\end{align}

A generalized risk-penalty regularization would be presented as follows:
\begin{align}
    \Omega (f)  =
    \begin{cases}
           \   (-\widehat{R}_{\mathrm{PAC}} (f)  )  ^t, \qquad \mathrm{if} \ \widehat{R}_{\mathrm{PAC} (f)  }<0,\\
            \quad \quad 0, \qquad \qquad \qquad \mathrm{otherwise,}
    \end{cases}
\end{align}
where $t\geq0$ is hyper-parameter and should be an integer.  Specially, when $t=1$ and $\lambda=1$, the formulation is the same as using the Rectified Linear Unit (ReLU) function as the correct function. When $t=1$ and $\lambda = 2$, the formulation is the same as using the absolute value (ABS) function as the correct function. Thus, this risk-penalty regularization could be regarded as a generalized method to deal with the overfitting problem caused by the negative empirical risk. And it could work well in our experiments. Then, the training objective with regularization term would be $\widehat{R}_{\mathrm{un}} (f)  +\lambda\Omega (f) $, where $\lambda$ is considered as the weight of the regularization term, and the  $\widehat{R}_{\mathrm{un}}$ is our proposed URE. We denote this regularized estimator as $PLLAC_{Reg}$.  The training procedure of PLLAC via $PLLAC_{Reg}$ is as Algorithm \ref{algo_disjdecomp}.

\IncMargin{1em}
\begin{algorithm}[tb]
  \SetKwData{Left}{left}\SetKwData{This}{this}\SetKwData{Up}{up}
  \SetKwFunction{Union}{Union}\SetKwFunction{FindCompress}{FindCompress}
  \SetKwInOut{Input} {Input}
  \Input{Classifier $f(\cdot)$, Iteration $T_{max}$, Epoch $I_{max}$, Parameter $\lambda,t$, Dataset $D=\{ (x_i,y_i)  \}^n_{i=1}$}
  \BlankLine
  \textbf{Initialize} Split dataset $D$ into training set and test set, generate partially labeled of each instance by generation model and obtain partial label training set $\tilde{D}={ (x_i,Y_i)  }_{i=1}^n$. And initialize $p (y_i=j|x_i)  =1$,$\forall j\in Y_i$, otherwise $p (y_i=j|x_i)  =0$;\\
  \For{$i\leftarrow 1$ \KwTo $T_{max}$}{
  \textbf{Shuffle} $\tilde{D}={ (x_i,Y_i)  }_{i=1}^n$.\\
    \For{$j\leftarrow 1$ \KwTo $I_{max}$}{
    \label{forins}
    \lIf{$\widehat{R}_{\mathrm{PAC}}<0$}{calculate $\lambda\Omega (f)  $, \textbf{update} model $f$ by $\widehat{R}_{\mathrm{un}}+\lambda\Omega (f)  $}
    \lElse{\textbf{update} model $f$ by $\widehat{R}_{\mathrm{un}}$}
    \textbf{Update} $p (y_i|x_i)  $;
    }
  }
  \SetKwInOut{Output}{Output}
  \Output{Model $f$}
  \caption{Training Algorithm of PLLAC}\label{algo_disjdecomp}
\end{algorithm}\DecMargin{1em}

\section{EXPERIMENTS}\label{sec:exp}
\subsection{Experimental Setup}

\begin{table}[ht]
\vspace{-1.0em}
    \centering
    \caption{Information of benchmark dataset and UCI dataset}
    \label{dataset1}
    \begin{tabular}{l|c|c|l|c|c}
    \toprule  
        benchmark dataset & \#size & model & UCI dataset &\#size & model\\ 
    \midrule
        MNIST & 70,000 & layers-MLP &har & 10,299  &linear model   \\ 
        Kuzushiji-MNIST & 70,000  & layers-MLP & msplice & 3,175  &linear model   \\  
        Fashion-MNIST & 70,000 & layers-MLP & optdigits & 5,620  &linear model \\ 
        SVHN & 99,289  & ResNet-32 &texture  & 5,500    &linear model \\ 
        CIFAR-10 & 60,000 & ResNet-32& usps  & 9,298     &linear model\\ 
        
        \bottomrule
    \end{tabular}
\end{table}
\begin{table}[ht]
    \centering
    \caption{Information of real-world datasets}
    \label{dataset2}
    \begin{tabular}{l|c|c|c|c|c}
    \toprule
        dataset & \#size& model  &  dataset & \#size & model \\ 
    \midrule
        
        Lost & 1,122  &linear model   &   Soccer Player & 17,472 &linear model  \\ 
        BirdSong & 4,998 &linear model   &    Yahoo! News & 22,991 &linear model \\ 
        \bottomrule
    \end{tabular}
    
\end{table}

\noindent\textbf{Datasets.} We conduct experiments on three types of datasets, i.e., six widely used benchmark datasets including MNIST\footnote{
http://yann.lecun.com/exdb/mnist/} \cite{lecun1998gradient}, Kuzushiji-MNIST\footnote{https://github.com/rois-codh/kmnist} \cite{clanuwat2018deep}, Fashion-MNIST\footnote{https://github.com/zalandoresearch/fashion-mnist} \cite{xiao2017fashion},  SVHN\footnote{http://ufldl.stanford.edu/housenumbers/} \cite{netzer2011reading}, CIFAR-10\footnote{
https://www.cs.toronto.edu/$\sim$kriz/cifar.html} \cite{krizhevsky2009learning} and CIFAR-100\footnote{
https://www.cs.toronto.edu/$\sim$kriz/cifar.html} \cite{krizhevsky2009learning}, five datasets from the UCI Machine Learning Repository\footnote{https://archive.ics.uci.edu} \cite{asuncion2007uci} including har, msplice, optdigits, texture and usps, and five datasets from real-world partial-label datasets\footnote{http://palm.seu.edu.cn/zhangml/Resources.htm\#partial\_data} including Lost \cite{cour2011learning}, MSRCv2 \cite{liu2012conditional}, BirdSong \cite{briggs2012rank}, Soccer Player \cite{zeng2013learning} and Yahoo! News \cite{guillaumin2010multiple}. The statistics of these datasets are listed in the Table \ref{dataset1} and Table \ref{dataset2}. To generate candidate label sets for benchmark datasets and UCI datasets, we adopt a uniform generation process, which assumes each partially labeled instance is independently drawn from a probability distribution with the following distribution:
\begin{align}
    \tilde{p} (\boldsymbol{x},Y)  =\sum_{i=1}^{k}p (Y|y=i)  p (\boldsymbol{x},y=i)  , \ \mathrm{where}\  p (Y|y=i)  =
    \begin{cases}
        \ \frac{1}{2^{k-1}-1}  \quad &\mathrm{if}\  i \in \mathcal{Y},\\
        \ \quad 0 
        &\mathrm{if}\  i \notin \mathcal{Y},
    \end{cases}
\end{align}
In the generation process, we assume the candidate label set $Y$ is independent of the instance $\boldsymbol{x}$ when its ground-truth label is given, i.e., $p (Y|\boldsymbol{x},y)  =P (Y|y)  $. In addition, we randomly split the UCI datasets and real-world datasets into a training set and test set in the ratio 80\% : 20\%.

For most datasets, we select one class as class $ac$. To ensure that the  class $ac$ never appears during training,  we remove a sample from training set and add it into test set if the sample is annotated with class $ac$ in its candidate label set. Particularly, we select 54 classes out of the 171 classes in Soccer Player and regard all of them as augmented classes. Overall, about 20\%$\sim$30\% of the training data in each dataset are removed because they are labeled with class $ac$.\\
\textbf{Metrics}. We choose three evaluation metrics: accuracy, Macro-F1 and AUC to evaluate our methods. Accuracy shows the proportion of test instances of which predicted results are true labels. Macro-F1 and AUC could consider the precision, recall, F1-score and ROC curve comprehensively when evaluating the capability and performance of models. Their definitions are as follows:
\begin{itemize}
    \item \textbf{Macro-F1:} The macro-averaged F1 score (or Macro-F1 score)   is computed using the arithmetic mean (unweighted mean)   of all the per-class F1 scores, which treats all classes equally regardless of their support values.
    \item \textbf{AUC:} AUC stands for "Area under the ROC Curve", which shows the trade-off between sensitivity  (or True Positive Rate, TPR)   and specificity (1-False Positive Rate, FPR)  . It measures the entire two-dimensional area underneath the entire ROC curve from  (0,0)   to  (1,1)   and provides an aggregate measure of performance across all possible classification thresholds.
\end{itemize}
\textbf{Compared Methods.} At present, there is no model for partial label learning with augmented classes specifically. We compare our PLLAC method with other five PLL methods. In order to adapt these models to the PLLAC problem, we first utilize partially labeled dataset to train the compared methods and obtain a k-class classifier, then set a threshold (0.95) for model's output to determine whether test sample belongs to \textit{ac}, that is, if the maximum value of the model's output does not exceed 0.95, the sample would be predicted to be class \textit{ac}.

The compared PLL methods are as follows:
\begin{itemize}
    \item[$\bullet$] RC \cite{RCCC}: a risk-consistent PLL method based on the importance of re-weighting strategy.
    \item[$\bullet$] CC \cite{RCCC}: a classifier-consistent PLL method based on the assumption that candidate label sets are generated uniformly.
    \item[$\bullet$] CAVL \cite{PLL4}: Based on RC, it improves the approach of confidence updating in combination with the Class Activation Mapping (CAM)  .
    \item[$\bullet$] PRODEN \cite{PRODEN}:  a progressive identification PLL method considering that only the true label contributes to retrieving the classifier and accomplishing classifier learning and label identification simultaneously.
    \item[$\bullet$] LWPLL \cite{LWS}: It provides a PLL loss that introduces the leverage parameter $\beta$ to consider the trade-off between losses on partial labels and non-partial ones. 
    \item[$\bullet$] VALEN \cite{PLL5}: an instance-dependent PLL method, which assumes that each instance is associated with a latent label distribution constituted by the real number of each label and recovered the label distribution as a label enhancement  (CE)   process in training.
\end{itemize}

Besides, we also compare with some complementary learning methods \cite{feng2020learning,feng2020learning,ishida2019complementary} for we can transform our partially labeled datasets into complementarily labeled dataset by regarding non-candidate labels as complementary labels. Loss functions used for learning with multiple complementary labels like bounded multi-class loss functions MAE (Mean Absolute Entropy), MSE (Mean Square Entropy), 
and the upper-bound surrogate loss function EXP \cite{feng2020learning} are employed in the derived empirical risk estimator. In PLLAC, the three loss functions would be:
\begin{align}
    \nonumber
    \mathcal{L}_{\mathrm{MAE}}(f(\boldsymbol{x}),S)&=\sum_{j=1}^{k}\left|p(y=j|\boldsymbol{x})-y_i\right| \\
    \nonumber
    \mathcal{L}_{\mathrm{MSE}}(f(\boldsymbol{x}),S)&=\sum_{j=1}^{k}(p(j|\boldsymbol{x})-y_j)^2\\
    \nonumber
    \mathcal{L}_{\mathrm{EXP}}(f({\boldsymbol{x}}),\bar{S})&=\exp(-\sum_{j\notin\bar{S}}p(j|\boldsymbol{x}))
\end{align}
where $p_{\theta}(y|\boldsymbol{x})=\exp(f_y(\boldsymbol{x}))/\sum_{j=1}^k\log(f_j(\boldsymbol{x}))$ denotes the predicted probability of the instance $\boldsymbol{x}$ belonging to class $y$, $\bar{S}$ denotes the non-candidate labels of instance $\boldsymbol{x}$.

\noindent\textbf{Implementation Details.} For the different complexity of the instance features in different datasets, we instantiate the backbone network with different network structures for them. As listed in Table \ref{dataset1}, we choose three-layer  ($d$-500-$k$) MLP and 34-layer ResNet as classifiers in MNIST datasets and SVHN datasets. Since the scales of UCI datasets are not large and the most existing PLL methods adopt linear model, we also choose linear model as backbone network for them.
We search learning rate and weight decay from $\{10^{-4},\dots, 10^{-2}\}$. We set the mini-batch size to 256, and set the number of total training epochs to 200 and 150 on real-world datasets and other datasets, respectively. Our code is available at \href{https://github.com/hujiayu1223/PLLAC-project}{https://github.com/hujiayu1223/PLLAC-project}.

For fair comparison, we employ the same dataset construction process, backbone network, batch size and the total number of training epochs on all the compared methods. The compared experiments are based on open-source code provided by the authors of the paper. All of them adopt Adam optimizer and a fixed learning rate. In addition, we set the parameter $lw0$ and $lw$ to 2 and 1 respectively in LWPLL method as weights of two types of sigmoid loss, which satisfies the condition that the leveraged parameter is 2. In VALEN experiment, we set $\alpha$ and $\gamma$ to 0.1 and 5 respectively as the balance parameters of the loss function. 

We run all the experiments for 5 trails on every dataset and report the mean accuracy with standard deviation  (mean $\pm$ std).


\begin{table}[t]
    \centering
    \caption{Test performance in accuracy (mean$\pm$std) of different partial label loss functions on three types of datasets.}  
\label{table:pll_loss}
  \scalebox{1.0}{\small
  \begin{tabular}{c|ccccc}
    \toprule
                          & MNIST & Kuzushiji-MNIST & Fashion-MNIST & SVHN & CIFAR-10 \\ \midrule
        $\mathrm{PLLAC}_{RC}$ & 0.961±0.001 & 0.817±0.004 & 0.845±0.003 & 0.904±0.006 & 0.625±0.007\\ 
          $\mathrm{PLLAC}_{PRODEN}$ & 0.960±0.001 & 0.814±0.004 & 0.844±0.001 & 0.320±0.011 & 0.789±0.008 \\ 
        $\mathrm{PLLAC}_{CC}$ & 0.686±0.009 & 0.533±0.008 & 0.644±0.008 & 0.114±0.007  & 0.410±0.045\\ 
        $\mathrm{PLLAC}_{LWPLL}$ &0.532±0.051&0.530±0.049&0.338±0.291&0.174±0.131&0.130±0.013 \\\hline
        & har & msplice & optdigits & texture & usps \\ \cline{1-6}
        $\mathrm{PLLAC}_{RC}$& 0.927±0.008 & 0.914±0.009 & 0.933±0.011 & 0.759±0.032 & 0.911±0.005\\ 
         $\mathrm{PLLAC}_{PRODEN}$  & 0.927±0.007  & 0.913±0.008  & 0.930±0.009 & 0.695±0.025 & 0.910±0.004 \\ 
         $\mathrm{PLLAC}_{CC}$ & 0.663±0.040 & 0.802±0.012 & 0.632±0.014 & 0.365±0.013 & 0.662±0.006\\ 
         $\mathrm{PLLAC}_{LWPLL}$ &0.226±0.077&0.459±0.006&0.421±0.263&0.091±0.000&0.757±0.031 \\\hline
         & Lost & MSRCv2 & BirdSong & Soccer Player & Yahoo! News \\ \cline{1-6}
        $\mathrm{PLLAC}_{RC}$& 0.568±0.007 & 0.346±0.019 & 0.582±0.012 & 0.527±0.006 & 0.494±0.003\\ 
           $\mathrm{PLLAC}_{PRODEN}$ & 0.563±0.010  & 0.343±0.020 & 0.514±0.014 & 0.489±0.008 & 0.598±0.004 \\ 
          $\mathrm{PLLAC}_{CC}$& 0.477±0.030 & 0.237±0.012  & 0.385±0.004  & 0.326±0.002 & 0.431±0.002\\ 
          $\mathrm{PLLAC}_{LWPLL}$ &0.606±0.021&0.306±0.008&0.538±0.032&0.509±0.005&0.488±0.003 \\
        \bottomrule 
    \end{tabular}}
\end{table}
\begin{table}[t]
    \centering
    \caption{Test performance in Macro F1 (mean$\pm$std) of different partial label loss functions on three types of datasets.}  
\label{table:pll_loss}
  \scalebox{1.0}{\small
  \begin{tabular}{c|ccccc}
    \toprule
                          & MNIST & Kuzushiji-MNIST & Fashion-MNIST & SVHN & CIFAR-10 \\ \midrule
        $\mathrm{PLLAC}_{RC}$ & 0.959±0.002 &0.811±0.008 & 0.840±0.01 & 0.599±0.034 & 0.884±0.02\\ 
          $\mathrm{PLLAC}_{PRODEN}$ & 0.958±0.004 & 0.810±0.008 & 0.840±0.008 & 0.740±0.015 & 0.282±0.013 \\ 
        $\mathrm{PLLAC}_{CC}$ &  0.737±0.015 &0.580±0.008& 0.684±0.012 & 0.415±0.052  & 0.048±0.013\\ 
        $\mathrm{PLLAC}_{LWPLL}$ &0.403±0.063&0.338±0.291&0.262±0.299&0.062±0.061&0.059±0.013 \\\hline
        & har & msplice & optdigits & texture & usps \\ \cline{1-6}
        $\mathrm{PLLAC}_{RC}$& 0.831±0.038 & 0.887±0.006 & 0.938±0.008 & 0.773±0.048 & 0.827±0.013 \\ 
         $\mathrm{PLLAC}_{PRODEN}$  & 0.927±0.006   & 0.905±0.009 & 0.930±0.009&0.638+0.037&0.902+0.005 \\ 
         $\mathrm{PLLAC}_{CC}$ & 0.669±0.046  & 0.770±0.015&0.689±0.013&0.345±0.011&0.689±0.006\\ 
         $\mathrm{PLLAC}_{LWPLL}$ &0.117±0.085&0.326±0.225&0.414±0.006&0.015±0.000&0.719±0.050\\\hline
         & Lost & MSRCv2 & BirdSong & Soccer Player & Yahoo! News \\ \cline{1-6}
        $\mathrm{PLLAC}_{RC}$&0.520±0.014 & 0.222±0.016 & 0.442±0.009 & 0.232±0.012 & 0.615±0.011\\ 
           $\mathrm{PLLAC}_{PRODEN}$ & 0.515±0.021&0.205±0.010&0.399±0.016&0.250±0.012&0.615±0.008 \\ 
          $\mathrm{PLLAC}_{CC}$& 0.445±0.025&0.117±0.016&0.232±0.015&0.235±0.010&0.547±0.009\\ 
          $\mathrm{PLLAC}_{LWPLL}$ &0.506±0.041&0.139±0.008&0.382±0.031&0.082±0.011&0.488±0.011 \\
        \bottomrule 
    \end{tabular}}
\end{table}

\begin{table}[t]
    \centering
    \caption{Test performance in AUC (mean$\pm$std) of different partial label loss functions on three types of datasets.}  
\label{table:pll_loss}
  \scalebox{1.0}{\small
  \begin{tabular}{c|ccccc}
    \toprule
                          & MNIST & Kuzushiji-MNIST & Fashion-MNIST & SVHN & CIFAR-10 \\ \midrule
        $\mathrm{PLLAC}_{RC}$ & 0.999±0.001 & 0.978±0.001 & 0.986±0.001 & 0.936±0.011 & 0.994±0.006\\ 
          $\mathrm{PLLAC}_{PRODEN}$ & 0.999±0.000&0.979±0.001&0.986±0.001&0.970±0.001&0.795±0.008 \\ 
        $\mathrm{PLLAC}_{CC}$ & 0.996±0.001&0.955±0.003&0.980±0.002&0.847±0.020&0.807±0.008\\ 
        $\mathrm{PLLAC}_{LWPLL}$ &0.876±0.023&0.841±0.024&0.661±0.189&0.578±0.067&0.640±0.018 \\\hline
        & har & msplice & optdigits & texture & usps \\ \cline{1-6}
        $\mathrm{PLLAC}_{RC}$& 0.996±0.001 & 0.984±0.003 & 0.997±0.001 & 0.994±0.001 & 0.989±0.002\\ 
         $\mathrm{PLLAC}_{PRODEN}$  & 0.997±0.000   &0.985±0.003& 0.996±0.001&0.991±0.001&0.990±0.001 \\ 
         $\mathrm{PLLAC}_{CC}$ &0.996±0.001&0.978±0.004&0.992±0.001&0.987±0.002&0.986±0.000\\ 
         $\mathrm{PLLAC}_{LWPLL}$ &0.604±0.082&0.907±0.014&0.812±0.134&0.498±0.046&0.973±0.007 \\\hline
         & Lost & MSRCv2 & BirdSong & Soccer Player & Yahoo! News \\ \cline{1-6}
        $\mathrm{PLLAC}_{RC}$& 0.890±0.014 & 0.803±0.012 & 0.890±0.016 & 0.830±0.005 & 0.972±0.004\\ 
           $\mathrm{PLLAC}_{PRODEN}$ & 0.891±0.014&0.777±0.012&0.853±0.006&0.841±0.004&0.972±0.004 \\ 
          $\mathrm{PLLAC}_{CC}$& 0.860±0.017&0.772±0.007&0.777±0.010&0.838±0.002&0.966±0.004\\ 
          $\mathrm{PLLAC}_{LWPLL}$ &0.902±0.009&0.742±0.009&0.840±0.019&0.772±0.005&0.948±0.004\\
        \bottomrule 
    \end{tabular}}
\end{table}
\subsection{Impact of Partial Label Learning Losses}\label{sec:loss}
As stated in Section \ref{sec:URE}, $\ell_{\text{PLL}}$  in the first term of the derived URE $\widehat{R}_{\text{un}}$ can be any partial label learning losses. In this section, we choose three partial label learning losses including RC \cite{RCCC}, CC \cite{RCCC}, PRODEN \cite{PRODEN} and LWPLL \cite{LWS}, to instantiate $\ell_{\text{PLL}}$ and investigate the impact of partial label learning losses on three metrics: accuracy, Macro F1 and AUC. As shown in Table \ref{table:pll_loss}, PLLAC equipped with RC and PRODEN losses achieves more than 90\% accuracy on MNIST and most UCI datasets, which shows the effectiveness of our PLLAC methods. Moreover,  RC and PRODEN are overall superior than CC, and LWPLL performs well on real-world datasets. However, RC could achieve better results in most datasets. Therefore, we choose RC as our partial label learning loss in our further experiments.

\subsection{Comparison Experiments}
\begin{table}[htb]
    \centering
    \caption{Test performance in accuracy, Macro F1 and AUC (mean±std) of each method on benchmark datasets, where ResNet and MLP are employed as backbone network on CIFAR-10 and other three datasets, respectively. (The best ones are bolded, the next best ones are underlined)}
   \scalebox{1.0}{\small
    \label{table1}
  \begin{tabular}{c|c|cccccc}
    \toprule
     &Datasets & MNIST & Kuzushiji-MNIST & Fashion-MNIST & SVHN & CIFAR-10 & CIFAR-100\\ \midrule
     
    \multirow{10}{*}{Accuracy}&
        $\mathrm{PLLAC}_{Reg}$ & \textbf{0.961±0.001} & \textbf{0.817±0.004} & \textbf{0.845±0.003} & \textbf{0.904±0.006} & \textbf{0.625±0.007} &{\textbf{0.335±0.009}}\\ \cline{2-8}
        &PRODEN & 0.904±0.004 & 0.711±0.005 & 0.740±0.003 & 0.817±0.016 & 0.423±0.008 &{0.265±0.057}\\ 
        &CAVL & 0.891±0.005 & 0.704±0.005 & 0.567±0.070 & 0.686±0.064 & 0.221±0.034&{0.189±0.006} \\ 
        &VALEN & 0.562±0.003 & 0.510±0.001 & 0.557±0.004 & 0.693±0.013 & 0.491±0.004&{\underline{0.286±0.013}} \\
        &LWPLL & 0.673±0.086 & 0.532±0.048 & 0.343±0.081 & 0.706±0.065 & 0.219±0.035&{0.097±0.002} \\ 
        &RC & \underline{0.906±0.007} & \underline{0.744±0.004} & 0.751±0.005 & 0.778±0.016 & 0.625±0.007 &{0.278±0.003}\\ 
        &CC & 0.898±0.005 & 0.738±0.010 & 0.762±0.006 & \underline{0.862±0.024} & 0.525±0.012 &{0.218±0.006}\\  \cline{2-8}
        &MAE & 0.883±0.003 & 0.702±0.029 & \underline{0.772±0.026} & 0.369±0.031 & \underline{0.791±0.044}&{0.141±0.005} \\ 
        &MSE & 0.130±0.004 & 0.153±0.003 & 0.114±0.001 & 0.107±0.002 & 0.073±0.001&{0.107±0.001 }\\ 
       & EXP & 0.277±0.012 & 0.254±0.004 & 0.241±0.006 & 0.201±0.014 & 0.104±0.009&{0.090±0.000} \\ 
        \midrule
         \multirow{10}{*}{Macro-F1}&$\mathrm{PLLAC}_{Reg}$ & \textbf{0.959±0.002} & \textbf{0.811±0.008} & \textbf{0.840±0.01} & \textbf{0.599±0.034} & \textbf{0.884±0.02} &{\underline{0.345±0.008}}\\ \cline{2-8}
&PRODEN & \underline{0.910±0.005} & 0.758±0.004 & 0.770±0.007 & 0.463±0.006 & 0.825±0.013&{\textbf{0.348±0.063}} \\ 
        &CAVL & 0.889±0.008 & 0.731±0.006 & 0.603±0.072 & 0.179±0.062 & 0.554±0.110&{0.112±0.005} \\ 
        &VALEN & 0.373±0.014 & 0.196±0.009 & 0.356±0.010 & 0.168±0.018 & 0.385±0.127&{0.136±0.011} \\ 
        &LWPLL & 0.609±0.119 & 0.475±0.065 & 0.228±0.085 & 0.104±0.099 & 0.592±0.117&{0.010±0.002} \\ 
                
        &RC & 0.907±0.008 & \underline{0.781±0.004} & \underline{0.777±0.007} & 0.488±0.006 & \underline{0.841±0.005}&{0.254±0.004} \\ 
        &CC & 0.898±0.006 & 0.767±0.006 & 0.759±0.003& \underline{0.533±0.014} & 0.543±0.013&{0.110±0.006} \\ \cline{2-8}
        &MAE & 0.841±0.003 & 0.676±0.029 & 0.737±0.035 & 0.327±0.046 & 0.704±0.076&{0.037±0.004} \\ 
        &MSE & 0.099±0.004 & 0.128±0.003 & 0.050±0.001 & 0.031±0.003 &  0.025±0.001&{0.016±0.001} \\ 
        &EXP & 0.318±0.015 & 0.282±0.004 & 0.265±0.008 &  0.186±0.017  & 0.081±0.014 &{0.001±0.000}\\ 
        \midrule 
         \multirow{10}{*}{AUC}&$\mathrm{PLLAC}_{Reg}$ & \textbf{0.999±0.001} & \textbf{0.978±0.001} & \textbf{0.986±0.001} & \textbf{0.936±0.011} & \textbf{0.994±0.006} &{\textbf{0.929±0.003}}\\ \cline{2-8}
         &PRODEN & 0.922±0.003 & 0.765±0.004 & 0.804±0.008 & 0.578±0.003 & 0.833±0.015 &{0.307±0.019}\\ 
        &CAVL & 0.931±0.004 & 0.807±0.004 & 0.679±0.05 & 0.510±0.009 & 0.776±0.026&{0.339±0.007} \\ 
        &VALEN & 0.443±0.015 & 0.200±0.013 & 0.406±0.009 & 0.142±0.029 & 0.430±0.135&{0.044±0.009}\\ 
        &LWPLL & 0.851±0.034 & 0.774±0.021 & 0.692±0.063 & 0.505±0.022 & 0.773±0.033&{0.302±0.006} \\ 
        &RC & \underline{0.931±0.005} & 0.806±0.003 & 0.820±0.009 & 0.596±0.004 & 0.868±0.006 &{\underline{0.535±0.001}}\\ 
        &CC & 0.927±0.004 & 0.808±0.010 &  0.822±0.003 & \underline{0.674±0.011} & 0.682±0.013 &{0.290±0.014}\\ \cline{2-8}
        &MAE &  0.913±0.003 & \underline{0.819±0.029} & \underline{0.867±0.035} & 0.621±0.046 & \underline{0.869±0.076}&{0.282±0.015} \\
        &MSE &  0.482±0.004 & 0.489±0.003 & 0.487±0.001 &  0.488±0.003 &  0.496±0.001&{0.276±0.014} \\
        &EXP &  0.516±0.015 & 0.512±0.004 & 0.504±0.008 & 0.500±0.017 & 0.499±0.014&{0.275±0.014} \\ 
        \bottomrule
    \end{tabular}}
\end{table}

\begin{table}[htbp]
    \centering
    \caption{Test performance in accuracy, Macro F1 and AUC (mean±std) of each method on UCI datasets, where Linear is employed as backbone network. (The best ones are bolded, the next best ones are underlined)}
  \scalebox{1.0}{\small
  \label{table2}
  \begin{tabular}{c|c|ccccc}
    \toprule
    &datasets& har & msplice & optdigits & texture & usps \\ \midrule
         \multirow{10}{*}{Accuracy}& $\mathrm{PLLAC}_{Reg}$  & \textbf{0.927±0.008} & \textbf{0.914±0.009} & \textbf{0.933±0.011} & \textbf{0.759±0.032} & \textbf{0.911±0.005} \\ \cline{2-7}
         &PRODEN & 0.508±0.025 & 0.579±0.014 & 0.687±0.023 & 0.091±0.001 & 0.666±0.022 \\ 
       
        &CAVL & 0.432±0.132 & 0.424±0.042 & 0.727±0.065 & 0.091±0.000 & 0.571±0.085 \\ 
        &VALEN & 0.581±0.016 & 0.557±0.012 & 0.597±0.006 & 0.500±0.000 & 0.571±0.003 \\ 
        & LWPLL & 0.530±0.001 & 0.682±0.021 & 0.577±0.11 & 0.145±0.020 & 0.686±0.040 \\ 
        &RC & 0.527±0.025 & 0.580±0.014 & 0.782±0.016 & 0.093±0.002 & 0.690±0.018 \\
        &CC & 0.532±0.025 & 0.580±0.014 & \underline{0.805±0.017} & \underline{0.440±0.020} & 0.694±0.019 \\ \cline{2-7}
        &MAE & \underline{0.722±0.005} & \underline{0.702±0.021} & 0.683±0.120 & 0.141±0.039 & \underline{0.733±0.026} \\
        &MSE & 0.179±0.008 & 0.553±0.007 & 0.125±0.012 & 0.091±0.000 & 0.171±0.002 \\ 
        &EXP & 0.349±0.018 & 0.660±0.020 & 0.426±0.029 & 0.137±0.014 & 0.327±0.028 \\ 
        \midrule
        \multirow{10}{*}{Macro-F1}&
       $\mathrm{PLLAC}_{Reg}$  & \textbf{0.831±0.038} & \textbf{0.887±0.006} & \textbf{0.938±0.008} & \textbf{0.773±0.048} & \textbf{0.827±0.013} \\ \cline{2-7}
        &PRODEN & 0.518±0.029 & 0.409±0.038 & 0.736±0.019 & 0.016±0.002 & 0.699±0.02 \\ 
        &CAVL & 0.421±0.158 & 0.284±0.034 & 0.746±0.078 & 0.015±0.000 & 0.543±0.112 \\ 
        &VALEN & 0.422±0.034 & 0.584±0.014 & 0.447±0.014 & 0.091±0.000 & 0.436±0.009 \\ 
        &LWPLL & 0.453±0.005 & 0.635±0.028 & 0.496±0.126 & 0.081±0.017 & 0.672±0.055 \\ 
        &RC & 0.539±0.029 & 0.410±0.038 & 0.817±0.012 & 0.019±0.004 & 0.724±0.016 \\ 
        &CC & 0.544±0.029 & 0.411±0.038 & \underline{0.835±0.013} & \underline{0.453±0.021} & 0.728±0.017 \\ \cline{2-7}
        &MAE & \underline{0.671±0.003} & \underline{0.703±0.019} &  0.650±0.138  &  0.086±0.039 & \underline{0.751±0.041} \\ 
        &MSE & 0.068±0.010  & 0.337±0.017 & 0.064±0.018  & 0.015±0.000 & 0.039±0.005 \\ 
        &EXP & 0.347±0.021 & 0.609±0.022 & 0.484±0.034 & 0.092±0.017 & 0.317±0.033 \\ 
        \midrule
        \multirow{10}{*}{AUC}&$\mathrm{PLLAC}_{Reg}$ & \textbf{0.996±0.001} & \textbf{0.984±0.003} & \textbf{0.997±0.001} & \textbf{0.994±0.001} & \textbf{0.989±0.002} \\ \cline{2-7}
        &PRODEN & 0.622±0.013 & 0.554±0.014 & 0.736±0.016 & 0.500±0.000 & 0.703±0.017 \\ 
        &CAVL & 0.592±0.057 & 0.514±0.028 & 0.768±0.051 & 0.500±0.000 & 0.634±0.054 \\ 
        &VALEN & 0.423±0.041 & 0.414±0.029 & 0.513±0.014 & 0.015±0.000 & 0.447±0.012 \\ 
        &LWPLL & 0.722±0.023 & 0.667±0.019 & 0.739±0.066 & 0.505±0.002 & 0.752±0.036 \\ 
       & RC & 0.633±0.014 & 0.554±0.014 & 0.810±0.013 & 0.500±0.000 & 0.723±0.013 \\
        &CC & 0.636±0.014 & 0.554±0.013 & \underline{0.830±0.015} & \underline{0.588±0.009} & 0.726±0.014 \\ \cline{2-7}
        &MAE &  \underline{0.804±0.003} & \underline{0.747±0.019} & 0.776±0.138 & 0.506±0.039 & \underline{0.780±0.041} \\ 
        &MSE &  0.502±0.010 & 0.529±0.017 &  0.503±0.018 & 0.500±0.000 & 0.501±0.005 \\ 
        &EXP & 0.554±0.021 & 0.649±0.022 & 0.586±0.034 & 0.503±0.017 & 0.541±0.033 \\ 
        \bottomrule
    \end{tabular}}
\end{table}

\begin{table}[htbp]
    \centering
    \caption{Test performance in accuracy, Macro F1 and AUC (mean±std) of each method on real-world datasets, where Linear is employed as backbone network. (The best ones are bolded, the next best ones are underlined)}
  \scalebox{1.0}{\small
  \label{table3}
  \begin{tabular}{c|c|ccccc}
    \toprule
   &datasets& Lost &MSRCv2& BirdSong & Soccer Player & Yahoo! News \\ \midrule
       \multirow{10}{*}{Accuracy}&  $\mathrm{PLLAC}_{Reg}$ & \textbf{0.568±0.007} & \underline{0.346±0.019} & \textbf{0.582±0.012} & \textbf{0.527±0.006} & \textbf{0.494±0.003} \\ \cline{2-7}
       &PRODEN & 0.417±0.023 & 0.222±0.012 & 0.366±0.012 & 0.496±0.059 & 0.330±0.005 \\ 
       
        &CAVL & 0.387±0.023 & 0.236±0.006 & 0.308±0.012 & 0.405±0.021 & 0.336±0.004 \\ 
        &VALEN & 0.541±0.009 & \textbf{0.493±0.006} & \underline{0.481±0.004} & 0.482±0.004 & \underline{0.480±0.004} \\ 
         &LWPLL & \underline{0.559±0.031} & 0.285±0.009 & 0.351±0.010 & 0.479±0.003 & 0.227±0.004 \\ 
        &RC & 0.465±0.008 & 0.258±0.017 & 0.366±0.013 & 0.437±0.056 & 0.344±0.003 \\ 
        &CC & 0.416±0.002 & 0.293±0.017 & 0.367±0.012 & 0.473±0.010 & 0.337±0.004 \\ \cline{2-7}
        &MAE & 0.474±0.026 & 0.319±0.005 & 0.477±0.019 & \underline{0.516±0.009} & 0.449±0.011 \\ 
        &MSE & 0.264±0.032 & 0.149±0.012 & 0.263±0.013 & 0.402±0.011 & 0.368±0.003 \\ 
        &EXP & 0.194±0.011 & 0.105±0.002 & 0.239±0.011 & 0.173±0.006 & 0.206±0.001 \\ 
        \midrule
       \multirow{10}{*}{Macro-F1}&  $\mathrm{PLLAC}_{Reg}$ & \textbf{0.520±0.014} & \textbf{0.222±0.016} & \textbf{0.442±0.009} & \textbf{0.232±0.012} & \textbf{0.615±0.011} \\ \cline{2-7}
        &PRODEN & 0.350±0.021 & 0.124±0.016 & 0.265±0.012 & 0.098±0.007 & 0.150±0.012 \\ 
        &CAVL & 0.319±0.032 & 0.144±0.010 & 0.162±0.010 & 0.163±0.006 & 0.156±0.011 \\ 
        &VALEN & 0.404±0.030 & \underline{0.203±0.006} & 0.273±0.008 & \underline{0.425±0.009} & 0.002±0.003 \\ 
        &LWPLL & \underline{0.440±0.035} & 0.110±0.015 & 0.168±0.038 & 0.020±0.004 & 0.008±0.001 \\ 
       & RC & 0.397±0.013 & 0.157±0.016 & 0.264±0.016 & 0.125±0.008 & 0.231±0.010 \\ 
       & CC & 0.369±0.037 & 0.132±0.023 & 0.265±0.015 & 0.101±0.004 &  0.232±0.011\\ \cline{2-7}
       & MAE & 0.421±0.043 & 0.190±0.011  & \underline{0.371±0.026} & 0.170±0.009 & \underline{0.403±0.006} \\ 
       & MSE & 0.200±0.013  & 0.091±0.006 & 0.083±0.007  & 0.119±0.005 & 0.244±0.015 \\ 
       & EXP & 0.277±0.021 & 0.159±0.012 & 0.204±0.029  & 0.140±0.007 & 0.365±0.013 \\ 
        \midrule 
        \multirow{10}{*}{AUC}& $\mathrm{PLLAC}_{Reg}$ & \textbf{0.890±0.014} & \textbf{0.803±0.012} & \textbf{0.890±0.016} & \textbf{0.830±0.005} & \textbf{0.972±0.004} \\ \cline{2-7}
        &PRODEN & 0.546±0.016 & 0.495±0.011 & 0.503±0.012 & 0.503±0.007 & 0.483±0.006 \\ 
       & CAVL & 0.556±0.014 & 0.505±0.011 & 0.483±0.007 & 0.482±0.010 & 0.484±0.005 \\ 
        &VALEN & 0.318±0.027 & 0.110±0.010 & 0.128±0.004 & 0.063±0.004 & 0.136±0.006 \\ 
        &LWPLL & 0.593±0.010 & 0.510±0.013 & 0.464±0.013 & 0.549±0.007 & 0.472±0.001 \\ 
        &RC & 0.571±0.017 & 0.500±0.013 & 0.503±0.016 & 0.492±0.010 & 0.507±0.010 \\ 
        &CC & 0.556±0.024 & 0.494±0.010 & 0.503±0.015 & 0.507±0.006 & 0.508±0.011 \\ \cline{2-7}
        &MAE & \underline{0.596±0.043} & \underline{0.589±0.011} & \underline{0.561±0.026} &  \underline{0.648±0.009} & \underline{0.588±0.006} \\ 
        &MSE & 0.522±0.013 & 0.503±0.006 &  0.465±0.007 & 0.530±0.005 &  0.507±0.015 \\ 
        &EXP & 0.557±0.021 & 0.525±0.012 & 0.466±0.029 & 0.591±0.007 & 0.568±0.013 \\ 
        \bottomrule
    \end{tabular}}
\end{table}
Table \ref{table1}, \ref{table2} report the results of comparison experiments on benchmark datasets, Table \ref{table3} reports the results of that on real-world dataset. We could conclude our observations as follows:

First of all, whether on benchmark datasets or real-world datasets, our proposed method significantly outperforms the other compared methods in three different metrics, which fully demonstrates the effectiveness of $PLLAC_{\mathrm{Reg}}$ method in solving PLLAC problems. This may be because $PLLAC_{\mathrm{Reg}}$ method makes full use of unlabeled data which includes class  in the training stage, and could implicitly learn the distribution of augmented class from unlabeled data and partial labeled data, which helps to accurately identify augmented classes in the test set.

Meanwhile, we found that though some comparison methods can solve the PLLAC problem to a certain degree through the heuristic classification threshold setting, their performance varies greatly in different datasets and is not stable enough. This may due to the feature discrimination of instances from different classes is different with datasets varying. For example, if the features of instances in class $ac$  is highly similar to those of instances in class $kc$, it may be wrongly classified into the known class with a high probability, resulting in prediction failure. The heuristic threshold setting method is difficult to flexibly adapt to different datasets, which further reflects the necessity of designing models for PLLAC problems.

In addition, we find that the instance-dependent PLL model VALEN performs better on the real-world datasets than on the benchmark datasets. This is because the benchmark PLL dataset adopts a uniform partial label generation process, which is different from the assumption of VALEN. In real-world datasets, the partial labels is more likely to be instance-dependent, thus VALEN performs better.

Finally, we find that MAE, a complementary learning method, is a strong competitor, even surpassing compared PLL methods on some datasets. The advantage of transforming PLL into complementary learning over direct PLL is that it can learn from a large number of accurate supervised signals, i.e., instances must not belong to their non-candidate labels. Most PLL methods try to identify the ground-truth label from the candidate label set but may suffer from error accumulation problem due to misidentification, while converting the partial label learning task into a complementary learning task avoids this problem, which is probably the main reason why MAE can achieve a comparable performance to PLL methods.

\subsection{Performance of increasing unlabeled instances}

\begin{figure}[tbp]
\includegraphics[scale=0.3]{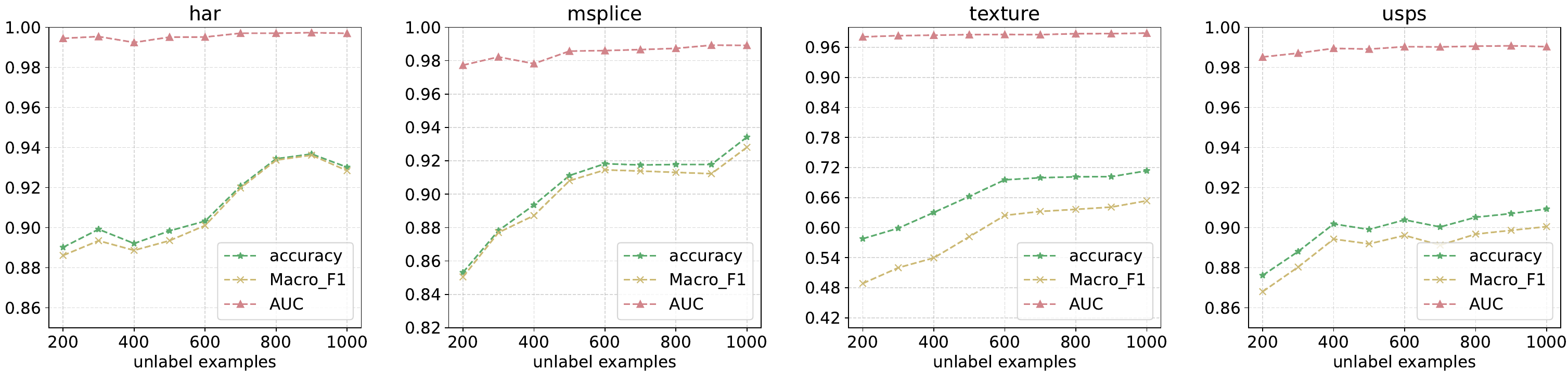}
\caption{Test performance on four UCI datasets when the number of unlabeled instances increases.}
\label{unlabel_num}
\end{figure}

The Theorem 1 in Section \ref{sec:theoretical}  claims that the performance of our proposed methods should be improved when more training instances are available. In this section, we verify this finding empirically by performing experiments on the UCI datasets.  It is natural for the classifier to get better as the number of partially labeled data increases, so we focus on the effect of the increase in unlabeled data on the performance. We keep the number of partially labeled data constant and vary the number of unlabeled instances from 200 to 2000. The results in Figure \ref{unlabel_num} show that when the number of unlabeled instances is increasing, the accuracy would increase first and then would gradually converge to an optimal value, which supports the derived error estimation bound in Theorem 1.

\begin{figure}[tbp]

\includegraphics[scale=0.34]{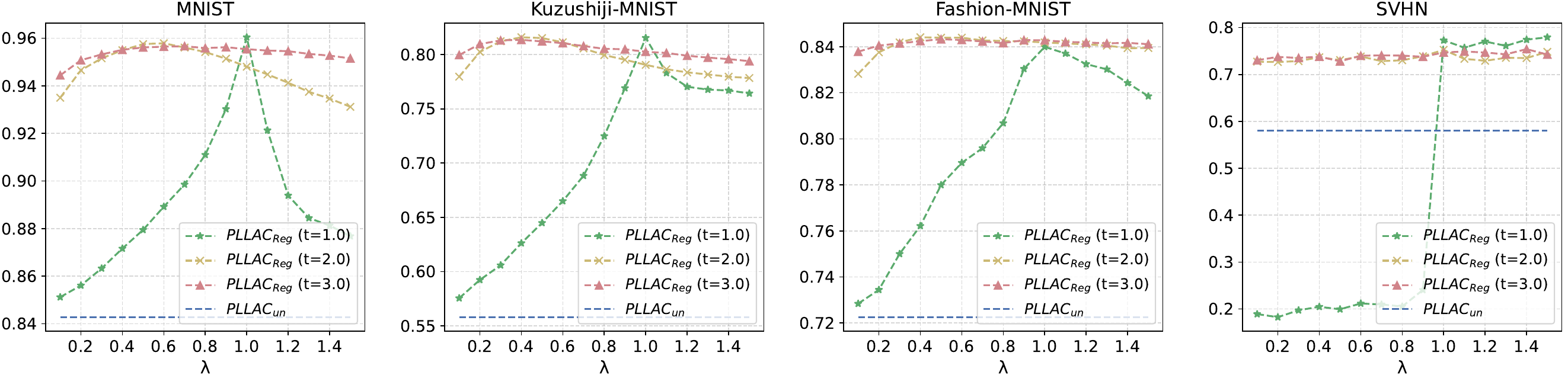}
\caption{Classification accuracy with different values of the regularization parameter $\lambda$ and $t$.}
\label{reg}
\end{figure}

\subsection{Analysis of Regularization parameter}
In this section, we first investigate the impact of risk-regularization by comparing the original PLLAC with a constructed model variant by removing the regularization term and  optimizing the unbiased risk estimator $\widehat{R}_{\mathrm{un}} (f)$ directly for model training, which is denoted $\mathrm{PLLAC}_{un}$. The results in Figure \ref{reg} show that $PLLAC_{un}$ performs worse than the $PLLAC_{Reg}$ regardless of $t=1$, $t=2$ and $t=3$, which indicates that the risk penalty regularization does alleviate the over-fitting problem caused by negative item in optimization objective.

Besides, we conduct parameter sensitivity analysis on the weighting of the risk-penalty regularization, i.e., $\lambda$, to investigate the effect of the risk-penalty regularization. We conduct experiments on four UCI datasets by varying $\lambda$ in $\{0.1,0.2,\dots,1.5\}$ and $t$ in $\{1.0,2.0,3.0\}$. As shown in Figure \ref{reg}, changing in $\lambda$ could make an improvement in accuracy first and degrade the performance after reaching the optimum. We find that the small $\lambda$ does not alleviate the over-fitting problem and causes NaN error during model training, which leads to terrible results, while a very large $\lambda$ makes the model focus more on the regularization term, which affects the optimization of the main loss for classification, and thus degrades the classification performance. The experimental results demonstrate the importance of the weights of risk-penalty regularization, i.e., $\lambda$.

\subsection{Influence of the mixture proportion}
To show the influence of the mixture proportion $\theta$, we conduct experiments on the Usps and Optdigits datasets by varying the preseted mixture proportion $\hat{\theta}$ from 0.1 to 1 under different values of the true mixture proportion $\theta$. As shown in Figure \ref{classshift} (a)-(b), performance improves as the estimated $\hat{\theta}$ approaches the true mixture proportion $\theta$, so it is important to estimate the true proportion accurately. Additionally, larger $\hat{\theta}$ could achieve better performance than smaller one in the case of inaccurate estimates.
\subsection{Handling Class Shift Condition}
\begin{figure}[tbp]
\centering
\includegraphics[scale=0.73]{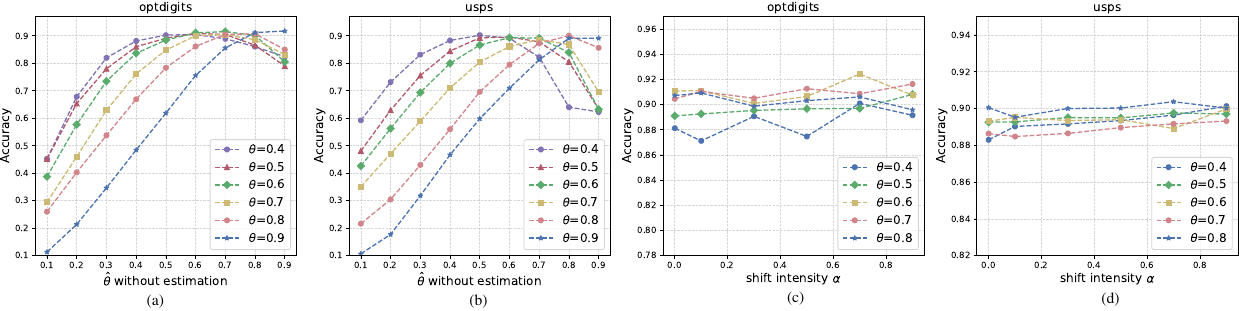}
\caption{(a)-(b): Influence of the mixture proportion $\theta$, (c)-(d): Sensitivity of PLLAC to class prior shift under different mixture proportion}
\label{classshift}
\end{figure}

To show our proposed method ability of handling more complex situation, we conduct experiments on Optdigits and Usps with class prior shift. Specifically, we select eight known classes and the rest is augmented classes, varying the preseted the mixture proportion $\theta$ in $\{0.4,0.5,0.6,0.7,0.8\}$, which means the distribution proportion of known classes and augmented classes is set by it. Then we use $\alpha$, selected in $\{0,0.1,0.3,0.5,0.7,0.9\}$ to control the shift intensity and reset the prior of eight known classes to $\{1-\alpha,1-\frac{3\alpha}{4},1-\frac{\alpha}{2},1-\frac{\alpha}{4},1+\frac{\alpha}{4},1+\frac{\alpha}{2},1+\frac{3\alpha}{4},1+\alpha\}$ in test data and Figure \ref{classshift} (c)-(d) reports the accuracy for different mixture proportion with different $\alpha$. As shown in Figure \ref{classshift} (c)-(d), performance of our method would not fluctuate greatly when $\alpha$ changes. This observation suggests that our proposed method effectively handles changing learning environments and is robust to class shift conditions, meaning that its performance does not degrade when class prior shifts.

\section{CONCLUSION}
In this paper, we investigate the problem of partial label learning with augmented classes and propose an unbiased risk estimator for it. We derive an estimation error bound for our methods, which ensures the optimal parametric convergence rate. Besides, to alleviate the over-fitting issue caused by negative empirical risk, we add a risk-penalty regularization term. Extensive comparison experiments on datasets prove that our proposed method is superior to other comparison methods, which verifies its effectiveness. Our method paves the way for the study of PLLAC. In the future, we will study more complex settings, such as the LAC tasks in scenarios such as instance-dependent PLL and noisy partial label learning, and apply the proposed methods to real-world scenarios.

\begin{acks}
This work was supported by the Chongqing Science and Technology Bureau (CSTB2022TTAD-KPX0180).
\end{acks}










\bibliographystyle{ACM-Reference-Format}
\bibliography{
    augmented_classes,
    partial_label,
    tist
}

\appendix

\section{proof for Theorem 1.}
Our proof of the estimation error bound is based on \emph{Rademacher complexity}.
Recall that the unbiased risk estimator we derived is represented as follows:
\begin{align}
\nonumber
\widehat {R}_{\mathrm{un}} (f)   &= \theta\frac{1}{n}\sum\limits_{i=1}^n[\ell_{\mathrm{PLL}} (f (\boldsymbol{x}_{i})  ,S_{i})  -\ell (f (\boldsymbol{x}_{i})  ,\mathrm{ac})  ] + \frac{1}{m}\sum\limits_{i=1}^m[\ell (f (\boldsymbol{x})  ,\mathrm{ac})  ]
\end{align}

Let us further introduce

\begin{align}
\nonumber
\widehat {R}_{\mathrm{kac}} (f)   &= \theta\frac{1}{n}\sum\limits_{i=1}^n[\ell_{\mathrm{PLL}} (f (\boldsymbol{x}_{i})  ,Y_i)  -\ell (f (\boldsymbol{x}_{i})  ,\mathrm{ac})  ] \\
\nonumber
 &= \theta\frac{1}{n}\sum\limits_{i=1}^{n}[\frac{1}{2} \sum_{o=1}^k \frac{p (y_{i}=o \mid \boldsymbol{x}_{i})  }{\sum_{j \in Y_{i}} p (y_{i}=j \mid \boldsymbol{x}_{i})  }\ell (f (\boldsymbol{x}_{i})  ,o)  -\ell (f (\boldsymbol{x}_{i})  ,\mathrm{ac})  ]\\
 \nonumber
 \widehat {R}_{\mathrm{tac}} (f)   &= \frac{1}{m}\sum_{j=1}^m\ell (f (\boldsymbol{x}_{j})  ,\mathrm{ac})  \\
 \nonumber
 R_{\mathrm{kac}} (f)   & =\mathbb{E}_{ (\boldsymbol{x}, S)   \sim P_{\mathrm{kc}}}[\mathcal{L}_{PLL} (f (\boldsymbol{x})  , S)  -\mathcal{L} (f (\boldsymbol{x})  , \mathrm{ac})  ] \\
\nonumber
R_{\mathrm{tac}} (f)   & =\mathbb{E}_{\boldsymbol{x} \sim P_{\mathrm{te}}}[\mathcal{L} (f (\boldsymbol{x})  , \mathrm{ac})  ]
\end{align}

\textbf{Lemma 1.} \emph{Assume the loss function $\mathcal{L} (f (\boldsymbol{x})  ,y)  $ is $\rho$-Lipschitz with respect to $f (\boldsymbol{x})   (0<\rho<\infty)  $ for all $y \in \mathcal{Y}$.Then, the following inequality holds:}
\begin{align}
\nonumber
\widetilde{\mathfrak{R}}_n\left (\mathcal{G}_{\mathrm{1}}\right)  \leq \sqrt{2} \rho \sum_{y=1}^k \mathfrak{R}_n\left (\mathcal{F}_y\right)  
\end{align}
where
\begin{align}
\nonumber
\mathcal{G}_{\mathrm{1}}&=\left\{ (\boldsymbol{x}, Y)   \mapsto \frac{1}{2} \sum_{i=1}^k \frac{p (y=i \mid \boldsymbol{x})  }{\sum_{j \in Y} p (y=j \mid \boldsymbol{x})  } \mathcal{L} (f (\boldsymbol{x})  , i)   \mid f \in \mathcal{F}\right\}\\
\nonumber
\mathcal{F}_y & =\left\{f: \boldsymbol{x} \mapsto f_y (\boldsymbol{x})   \mid f \in \mathcal{F}\right\} \\
\nonumber
\mathfrak{R}_n\left (\mathcal{F}_y\right)   & =\mathbb{E}_{p (\boldsymbol{x})  } \mathbb{E}_{\boldsymbol{\sigma}}\left[\sup _{f \in \mathcal{F}_y} \frac{1}{n} \sum_{i=1}^n f\left (\boldsymbol{x}_i\right)  \right] .
\end{align}
{\emph{Proof.} We introduce $p_i (\boldsymbol{x})  =\frac{p (y=i \mid \boldsymbol{x})  }{\sum_{j \in Y} p (y=j \mid \boldsymbol{x})  }$ for each instance $ (\boldsymbol{x},Y)  $. And we have $0\leq p_i (\boldsymbol{x})  \leq1$,$\forall i \in[k]$ and $\sum_{i=1}^k p_i (\boldsymbol{x})  =1$ since $p_i (\boldsymbol{x})  =0$ if $i \notin Y$. Then we can obtain $\widetilde{\mathfrak{R}}_n (\mathcal{G}_{\mathrm{1}})  \leq\mathfrak{R}_n (\mathcal{L}\circ\mathcal{F})  $ where $\mathcal{L}\circ\mathcal{F}$ denotes $\{\mathcal{L}\circ f|f \in \mathcal{F}\}$ . Since $\mathcal{F}_y =\left\{f: \boldsymbol{x} \mapsto f_y (\boldsymbol{x})   \mid f \in \mathcal{F}\right\}$ and the loss function $\mathcal{L} (f (\boldsymbol{x})  ,y)  $ is $\rho$-Lipschitz with respect to $f (\boldsymbol{x})   (0<\rho<\infty)  $ for all $y \in \mathcal{Y}$, by the Rademacher vector contraction inequality, we have $\mathfrak{R}_n (\mathcal{L}\circ\mathcal{F})  \leq\sqrt{2}\rho\sum_{y=1}^{k+1}\mathfrak{R}_n (\mathcal{F}_y)  $. \qed}

\textbf{Lemma 2.} \emph{Assume the multi-class loss function $\mathcal{L} (f (\boldsymbol{x})  ,y)  $ is $\rho$-Lipschitz $ (0<\rho<\infty)  $ with respect to $f (\boldsymbol{x})  $ for all $y \in \mathcal{Y}$ and upper bounded by a constant $C_{\mathcal{L}}$, i.e.,$C_{\mathcal{L}}=sup_{\boldsymbol{x}\in\mathcal{X},y\in\mathcal{Y},f\in\mathcal{F}}\mathcal{L} (f (\boldsymbol{x},y))  $. Then, for any $\delta > 0$,with probability at least $1-\delta$,we have}
\begin{align}
\nonumber
    \mathrm{sup}_{f\in\mathcal{F}}\left|R_{\mathrm{kac}} (f)  -\widehat{R}_{\mathrm{kac}} (f)  \right|\leq4\sqrt{2}\rho (k+1)  \frac{C_{\mathcal{F}}}{\sqrt{n}}+3C_{\mathcal{L}}\sqrt{\frac{\mathrm{log}\frac{2}{\delta}}{2n}} 
\end{align}
{\emph{Proof.} For any sample $S=(x_1,x_2,...,x_n)$, we define $\phi(S)$ that for any sample $S$ by
\begin{align}
\nonumber
    \phi(S)=\mathrm{sup}_{f\in\mathcal{F}} (R_{\mathrm{kac}} (f)  -\widehat{R}_{\mathrm{kac}}^{'} (f) )
\end{align}
Let $S$ and $S^{'}$ be two instances differing by exactly one point, say $x_n$ in $S$ and ${x_n}^{'}$ in $S^{'}$. Then since the difference of suprema does not exceed the supremum of the difference, we have
\begin{align}
    \phi(S)-\phi(S^{'}) &\leq \mathrm{sup}_{f\in\mathcal{F}} (\widehat{R}_{\mathrm{kac}} (f)  -\widehat{R}_{\mathrm{kac}}^{'} (f)  ) \\
    &=\mathrm{sup}_{f\in\mathcal{F}}\frac{f(x_n)-f({x_n}^{'})}{n}\leq \frac{3C_{\mathcal{L}}}{n}
\end{align}
therefore, when an instance $x_i$ in $\widehat{R}_{\mathrm{kac}} (f)  $ is replaced by another arbitrary instance ${x_i}^{'}$, and then the change of $\mathrm{sup}_{f\in\mathcal{F}} (R_{\mathrm{kac}} (f)  -\widehat{R}_{\mathrm{kac}} (f)  )  $ is no greater than $\frac{3C_{\mathcal{L}}}{n}$. Then, by applying the Diarmid's inequality (McDiarmid 1989 \cite{mohri2018foundations}), for any $\delta>0$, with probability at least $1-\frac{\delta}{2}$,
\begin{align}
\nonumber
\mathrm{sup}_{f\in\mathcal{F}}\left (R_{\mathrm{kac}} (f)  -\widehat{R}_{\mathrm{kac}} (f)  \right)  \leq\mathbb{E}\left[\mathrm{sup}_{f\in\mathcal{F}} (R_{\mathrm{kac}} (f)  -\widehat{R}_{\mathrm{kac}}(f))  \right]+3C_{\mathcal{L}}\sqrt{\frac{\mathrm{log}\frac{2}{\delta}}{2n}}
\end{align}
We next bound the expectation of the right-hand side as follows:
\begin{align}
    \nonumber
    \mathbb{E}\left[\mathrm{sup}_{f\in\mathcal{F}} (R_{\mathrm{kac}} (f)  
     -\widehat{R}_{\mathrm{kac}}(f))\right]
     &=\mathbb{E}\left[\mathrm{sup}_{f\in\mathcal{F}} \mathbb{E}\left[(\widehat{R}_{\mathrm{kac}}^{'} (f)  -\widehat{R}_{\mathrm{kac}}(f))\right]  \right]\\
    \nonumber
    &\leq \mathbb{E}\left[\mathrm{sup}_{f\in\mathcal{F}} (\widehat{R}_{\mathrm{kac}}^{'} (f)  -\widehat{R}_{\mathrm{kac}}(f))\right] \\
    \nonumber
    &= \mathbb{E}\left[\mathrm{sup}_{g\in\mathcal{L}\circ f} \sum_{i=1}^{n}(\frac{1}{2}p_i({x_i}^{'})g({x_i}^{'})-g({x_i}^{'})-(\frac{1}{2}p_i(x_i)\cdot g(x_i)-g(x_i))\right] \\
    \nonumber
    &\leq \mathbb{E}_{\sigma}\left[\mathrm{sup}_{g\in\mathcal{L}\circ f} \sum_{i=1}^{n}(\sigma_i\frac{1}{2}p_i({x_i}^{'})g({x_i}^{'})-\sigma_i g({x_i}^{'})-(\frac{1}{2}p_i(x_i)\sigma_i g(x_i)-g(x_i))\right]\\
    \nonumber
    &\leq \mathbb{E}_{\sigma}\left[ \mathrm{sup}_{g\in\mathcal{L}\circ f}\sum_{i=1}^{n}\sigma_i\frac{1}{2}p_i(x_i)g(x_i)-\sigma_i g(x_i)\right]+\mathbb{E}_{\sigma}\left[ \mathrm{sup}_{g\in\mathcal{L}\circ f}\sum_{i=1}^{n}\sigma_i\frac{1}{2}p_i(x_i)g(x_i)-\sigma_i g(x_i)\right]\\
    \nonumber
    &=2\mathbb{E}_{\sigma}\left[\mathrm{sup}_{g\in\mathcal{L}\circ f}\sum_{i=1}^{n}\sigma_i\frac{1}{2}p_i({x_i}^{'})g({x_i}^{'})\right]+2\mathbb{E}_{\sigma}\left[\mathrm{sup}_{g\in\mathcal{L}\circ f}\sigma_i g(x_i)\right]\\
    \nonumber
    &=2\mathfrak{R}_n\left (\mathcal{G}_{1}\right)  +2\mathfrak{R}_n\left (\mathcal{L}\circ\mathcal{F}\right)  \leq4\mathfrak{R}_n\left (\mathcal{L}\circ\mathcal{F}\right)\\
\end{align}
Considering $\mathfrak{R}_n (\mathcal{F}_y)  \leq C_{\mathcal{F}}/\sqrt{n}$, we have for any $\delta>0$, with probability at least $1-\frac{\delta}{2}$,
\begin{align}
\nonumber
    \mathrm{sup}_{f\in\mathcal{F}}\left (R_{\mathrm{kac}} (f)  -\widehat{R}_{\mathrm{kac}} (f)  \right)  \leq4\sqrt{2}\rho (k+1)  \frac{C_{\mathcal{F}}}{\sqrt{n}}+3C_{\mathcal{L}}\sqrt{\frac{\mathrm{log}\frac{2}{\delta}}{2n}}
\end{align}
Taking into account the other side $\mathrm{sup}_{f\in\mathcal{F}} (\widehat{R}_{\mathrm{kac}} (f)  -R_{\mathrm{kac}} (f)  )  $, we have for any $\delta>0$, with probability at least $1-\delta$,
\begin{align}
  \nonumber
    \mathrm{sup}_{f\in\mathcal{F}}\left|R_{\mathrm{kac}} (f)  -\widehat{R}_{\mathrm{kac}} (f)  \right|\leq4\sqrt{2}\rho (k+1)  \frac{C_{\mathcal{F}}}{\sqrt{n}}+3C_{\mathcal{L}}\sqrt{\frac{\mathrm{log}\frac{2}{\delta}}{2n}} 
\end{align}
which concludes the proof.\qed}

\textbf{Lemma 3.} \emph{Assume the multi-class loss function $\mathcal{L} (f (\boldsymbol{x})  ,y)  $ is $\rho$-Lipschitz $ (0<\rho<\infty)  $ with respect to $f (\boldsymbol{x})  $ for all $y \in \mathcal{Y}$ and upper bounded by a constant $C_{\mathcal{L}}$, i.e.,$C_{\mathcal{L}}=sup_{\boldsymbol{x}\in\mathcal{X},y\in\mathcal{Y},f\in\mathcal{F}}\mathcal{L} (f (\boldsymbol{x},y)  $.Then,for any $\delta > 0$,with probability at least $1-\delta$,we have}
\begin{align}
\nonumber
    \mathrm{sup}_{f\in\mathcal{F}}\left|R_{\mathrm{tac}} (f)  -\widehat{R}_{\mathrm{tac}} (f)  \right|\leq2\sqrt{2}\rho (k+1)  \frac{C_{\mathcal{F}}}{\sqrt{m}}+C_{\mathcal{L}}\sqrt{\frac{\mathrm{log}\frac{2}{\delta}}{2m}} 
\end{align}
{\emph{Proof.} Lemma 3 can be proved as Lemma 2 at the same way.\qed}

\textbf{Lemma 4.} \emph{Let $\widehat{f}_{\mathrm{un}}$ be the empirical risk minimizer (i.e., $\widehat{f}_{\mathrm{un}}=\arg\min_{f\in\mathcal{F}}\widehat{R} (f)  $)   and $f^*$ be the true risk minimizer (i.e., $f^*=\arg\min_{f\in\mathcal{F}}R (f)  $)  , then the following inequality holds:}
\begin{align}
\nonumber
    R (\widehat{f})  -R (f^*)  \leq2\mathop{\mathrm{sup}}\limits_{f\in\mathcal{F}}\left|\widehat{R}_{\mathrm{un}} (f)  -R_{\mathrm{un}} (f)  \right|
\end{align}
{\emph{Proof.} It is intuitive to obtain that
\begin{align}
\nonumber
    R (\widehat{f})  -R (f^*)  &\leq R_{\mathrm{un}} (\widehat{f})  -\widehat{R}_{\mathrm{un}} (\widehat{f})  +\widehat{R}_{\mathrm{un}} (\widehat{f})  -R_{\mathrm{un}} (f^*)  \\
    \nonumber
    &\leq R_{\mathrm{un}} (\widehat{f})  -\widehat{R}_{\mathrm{un}} (\widehat{f})  +R_{\mathrm{un}} (\widehat{f})  -R_{\mathrm{un}} (f^*)  \\
    \nonumber
    &\leq2\mathop{\mathrm{sup}}\limits_{f\in\mathcal{F}}\left|\widehat{R}_{\mathrm{un}} (f)  -R_{\mathrm{un}} (f)  \right|
\end{align}
which completes the proof.

Combining Lemma 1, Lemma 2, Lemma 3, and Lemma 4, Theorem 1 is proved. \qed}
\end{document}